\documentclass[letter, 10pt, conference]{ieeeconf}
\IEEEoverridecommandlockouts

\usepackage{cite}

\usepackage{amsmath,amssymb,amsfonts}
\usepackage{algorithmic}
\usepackage{graphicx}
\usepackage{subcaption}
\usepackage{textcomp}
\usepackage{xcolor}
\usepackage{bm}
\usepackage{booktabs}
\usepackage{tikz}
\usepackage{hyperref}
\usepackage{epstopdf}

\def\BibTeX{{\rm B\kern-.05em{\sc i\kern-.025em b}\kern-.08em
    T\kern-.1667em\lower.7ex\hbox{E}\kern-.125emX}}
\usetikzlibrary{arrows,decorations.pathmorphing,backgrounds,fit,positioning,calc,shapes}

\begin{document}

\title{Combining Context Awareness and Planning \\ to Learn Behavior Trees from Demonstration
\thanks{
\textbf{This work has been submitted to the IEEE for possible publication. Copyright may be transferred without notice, after which this version may no longer be accessible.}
This project is financially supported by the Swedish Foundation for Strategic Research and by the Wallenberg AI, Autonomous Systems, and Software Program (WASP) funded by the Knut and Alice Wallenberg Foundation. The authors gratefully acknowledge this support.}
}

\author{\authorblockN{
Oscar Gustavsson\authorrefmark{1}$^{a}$,
Matteo Iovino\authorrefmark{1}$^{a,b}$,
Jonathan Styrud$^{a,c}$ and
Christian Smith$^{a}$}\\
\thanks{\authorrefmark{1} These authors have contributed equally to this work.}
\thanks{$^{a}$Division of Robotics, Perception and Learning, KTH - Royal Institute of Technology, Stockholm, Sweden}
\thanks{$^{b}$ABB Corporate Research, Västerås, Sweden}
\thanks{$^{c}$ABB Robotics, Västerås, Sweden}
}

\maketitle

\begin{abstract}
Fast changing tasks in unpredictable, collaborative environments are typical for medium-small companies, where robotised applications are increasing. Thus, robot programs should be generated in short time with small effort, and the robot able to react dynamically to the environment. To address this we propose a method that combines context awareness and planning to learn Behavior Trees (BTs), a reactive policy representation that is becoming more popular in robotics and has been used successfully in many collaborative scenarios. Context awareness allows to infer from the demonstration the frames in which actions are executed and to capture relevant aspects of the task, while a planner is used to automatically generate the BT from the sequence of actions from the demonstration. The learned BT is shown to solve non-trivial manipulation tasks where learning the context is fundamental to achieve the goal. Moreover, we collected non-expert demonstrations to study the performances of the algorithm in industrial scenarios.
\end{abstract}

\begin{keywords}
Behavior Trees, Learning from Demonstration, Manipulation, Collaborative Robotics
\end{keywords}

\section{Introduction}


The focus for robotized applications is shifting towards medium-small companies, where production is often customer specific and achieved in shorter cycles, resulting in frequent changes in the tasks of the robot. At the same time, robots are more frequently used collaboratively, sharing the environment with humans. Consequently, there is a need for methods for generating robot programs that do not require high programming skills or long time, and give programs that react to changes in the environment. This need can be fulfilled by using learning algorithms to synthesize control policies that are reactive, human-readable, and modular - allowing for code reusability. Behavior Trees (BTs) have proven to be a good policy representation for many robotic applications~\cite{colledanchise_behavior_2018} and the interest in methods for their automatic synthesis is also increasing~\cite{iovino_survey_2020}.

In this paper, we propose to automatically generate BTs from human demonstration. Learning from Demonstration (LfD) methods (also known as Programming by Demonstration or Imitation Learning) allow the generation of computer programs in an intuitive way, reducing the programming skills required by the user, while leveraging the user experience in solving specific tasks. When demonstrating a task, the user naturally performs actions in the correct sequence leading to the goal, thus task switching controllers like BTs can benefit from this, encoding task hierarchy in their representation. Vice-versa, by representing a policy as a BT, we inherit the reactivity that allows robots to adapt to changes in the environment.

Our contribution is a method that learns task constraints and contexts, to leverage the idea of backchaining to correctly design the BT: starting from the goal, pre-conditions are iteratively expanded with actions that achieve them - those actions that have that particular condition as their post-conditions. Then, those actions' unmet pre-conditions are expanded in the same way. Examples explaining the backchaining are provided later on. When compared to the state of the art, our method allows to learn only the relevant parts of a task, thus limiting the size of the BT. Moreover, our method learns the tree directly, thus not needing to resort to other policy representations as intermediate learning steps.

\begin{figure}[t]
    \centering
    \resizebox{\linewidth}{6.9cm}{
        \includegraphics{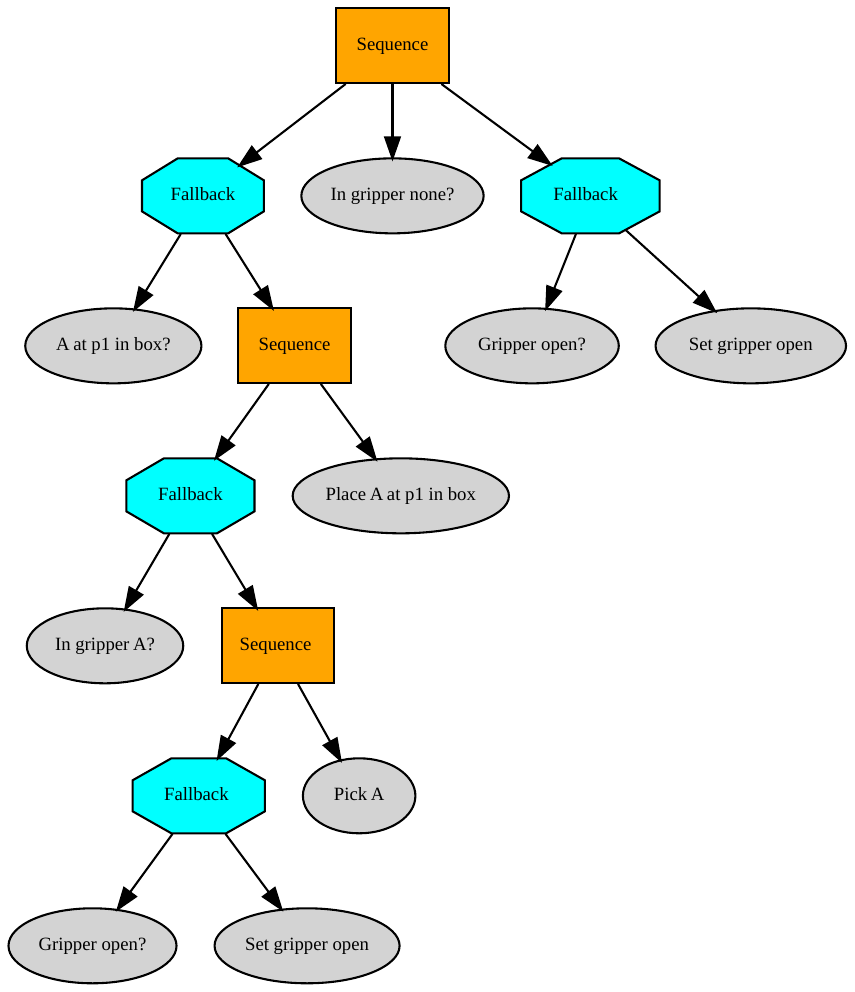}
    }
    \caption{Behavior Tree solving the task: object in a box.}
    \label{fig:example_bt}
\end{figure}

\section{Background and Related Work} \label{sec:work}

This section provides a background on Behavior Trees and Learning from Demonstration, summarizes related work, and shows how our method contributes to the state of the art.

\subsection{Behavior Trees}

Behavior Trees are task switching policy representations that originated in the gaming industry as an alternative architecture to Finite State Machines (FSM)~\cite{colledanchise_behavior_2018}. BTs have explicit support for task hierarchy, action sequencing and reactivity, and improve on FSMs especially in terms of modularity and reusability~\cite{iovino_survey_2020}. 
\par
In a BT, internal nodes are called \emph{control nodes}, (polygons in Figure~\ref{fig:example_bt}), while leaves are called \emph{execution nodes} or \emph{behaviors} (ovals). During execution, \textit{tick} signals are propagated from the root down the tree at a specified frequency. The most common types of control nodes are \emph{Sequence} and \emph{Fallback} (or \emph{Selector}). Sequence nodes execute their children in a sequence, returning once all succeed or one fails. Fallback nodes also execute their children in a sequence but return when one succeeds or all fail. Execution nodes execute a behavior when ticked and return one of the status signals \textit{Running}, \textit{Success} or \textit{Failure}. They are of type Action nodes or Condition nodes, the latter encoding status checks and sensory feedback and immediately returning \textit{Success} or \textit{Failure}. BTs are functionally close to decision trees with the main difference in the \textit{Running} state that allows BTs to execute actions for longer than one tick. The \textit{Running} state is also crucial for reactivity, allowing other actions to preempt non-finished ones. For more detail on BTs, see e.g.~\cite{colledanchise_behavior_2018}.

As is shown in the attached video\footnote{\url{https://youtu.be/cy6PKRrsMjM}}, when representing a policy learned from demonstration with a BT, we automatically benefit from the BT reactivity, as some actions in the demonstrated sequence can be skipped by the robot, if already performed by the user, or executed again if external disturbances undo some parts of the task.

\subsection{Learning from Demonstration}


In LfD, robot programs are learned from human demonstrations~\cite{ravichandar_recent_2020}. LfD methods are particularly useful when users don't have enough programming skills, or writing robot programs to solve a task takes too long. A LfD method defines how to demonstrate a task, the policy representation and how that policy is learned. Demonstration methods are mainly of three types: \textit{kinesthetic teaching}, where the user physically moves the robot, \textit{teleoperation}, where a robot is controlled through an external device - particularly useful when the robot operates in unreachable environments - and \textit{passive observation}, where the robot or the human are endowed with tracking systems and the demonstrator's body motion is recorded.
When choosing a demonstration method, attention has to be put in the correspondence problem, i.e. the mapping between a motion performed by a human teacher and the one executed by the robot.
Kinesthetic teaching, which is used in this paper, does not suffer from this problem as the motion is directly recorded in the robot task (or joint) space. This method is intuitive and no particular training is required for the user. However, there is a limit in the number of robot degrees of freedom and weight that a human can move to demonstrate a task.

\subsection{Related Work}


To the best of our knowledge there are only four previous studies~\cite{sagredo-olivenza_trained_2019, french_learning_2019, robertson_building_2015, suddrey_learning_2021} that combine LfD and BTs. The method proposed in~\cite{sagredo-olivenza_trained_2019} and ~\cite{french_learning_2019} learns a mapping from state space to action space as a Decision Tree (DT), which is converted into a BT, using the fact that BTs generalize DTs \cite{colledanchise_how_2017}. In~\cite{robertson_building_2015}, on the other hand, authors encode the demonstrated sequence of actions directly as a BT, while in \cite{suddrey_learning_2021} BTs are generated by natural language instructions.

In~\cite{sagredo-olivenza_trained_2019}, LfD is used to assist game designers in creating behaviors for Non-Player Characters (NPCs). A DT is trained as a mapping between the game's state and the action the NPC should execute. The DT is later flattened into a set of rules which are simplified with a greedy algorithm and translated into a BT. The final BT, however, cannot be used as is but requires a final tuning of parameters, thus limiting its usage to support the final NPC design. This is extended by~\cite{french_learning_2019} who generalize it to use any logic minimizer instead of the greedy algorithm, and implement the solution on a mobile manipulator performing a house cleaning task. The whole action space and state space are encoded in the tree which would then contain elements that are not used at runtime and complicate the structure of the BT. Furthermore, the frame of reference is hard-coded in the actions which reduces their reusability. The same \emph{place} action for example cannot be reused in different situations. The method does not exploit any previous knowledge about the actions' behaviors and it cannot execute behaviors that were not demonstrated but that may be required in some situations. 

Authors of~\cite{robertson_building_2015} synthesize a BT directly. They train an agent to play the video game StarCraft from expert demonstrations. Each demonstration consists of a sequence of actions that are placed under a sequence node in the BT, and all sequence nodes for each demonstration are placed under a fallback node. The BT is finally simplified by finding similarities between different demonstrations. We argue that this approach is more of a mapping between in-game actions and BT behaviors, resulting in large and hard to read BTs ($>50.000$ nodes) and limiting the reactivity (in-game actions and conditions are all put under the same Sequence node).

Finally, in~\cite{suddrey_learning_2021}, authors propose a method able to generate BTs out of natural language instructions. The method parses the expression and searches in the robot database if there are already trees solving the requested task. Otherwise, a new tree is learned by matching the parsed expression to hand-coded primitive methods encoding simple BTs solving e.g. the picking of an object. Although the method allows to define a wide variety of tasks, the learned trees are hard to read and it is not clear how would the method handle assembly-like tasks where the relative position of objects in the scene matters.

Not using BTs, there are other LfD methods that have been used to learn a task plan. Some authors have focused on learning some kind of FSM. In~\cite{niekum_learning_2015} an FSM is learned while in~\cite{hovland_skill_1996} a Hidden Markov Model (HMM) is learned. A different approach is taken by the authors in~\cite{konidaris_robot_2012} where skill chains are created by chaining skills that have a goal that allows the next skill to be executed successfully. Chains from multiple demonstrations are combined into a skill tree to allow multiple chains that achieve the same goal. The main advantage of using BTs over these methods is the increased readability and inherent reactivity of BTs.

The method we propose in this paper builds on the work of~\cite{ekvall_robot_2008} and~\cite{colledanchise_towards_2019}. Authors in~\cite{ekvall_robot_2008} use a demonstration to infer constraints about the order in which actions should be performed, and, if multiple demonstrations are provided, the robot can learn to generalize, instead of just repeating what the humans do. When the actions are given pre- and post-conditions, a planner can be used to construct a plan that satisfies the demonstrated constraints, leveraging the idea of backchaining. \emph{Pre-conditions} are the conditions that must be satisfied prior to executing the action, while \emph{post-conditions} are the conditions that will be satisfied as an effect of the action execution. For example, taking Table~\ref{tab:actions} as reference, a pre-condition to the \textit{Pick} action is that the robot gripper is open, while post-conditions are that the gripper is closed and an object is held.
Colledanchise et al.~\cite{colledanchise_towards_2019} present a planner to automatically grow a BT. The algorithm grows the tree iteratively, where in each iteration the current tree is executed and failing pre-conditions are replaced with subtrees that execute actions with appropriate post-conditions. 

\begin{figure}[tbp]
    \centering
    \subcaptionbox{Place to the left}[0.49\linewidth]{
        \centering
        \includegraphics[scale=0.3]{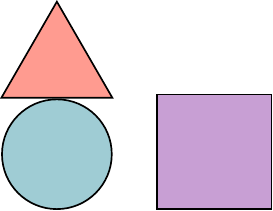}
    }
    \subcaptionbox{Place to the right}[0.49\linewidth]{
        \centering
        \includegraphics[scale=0.3]{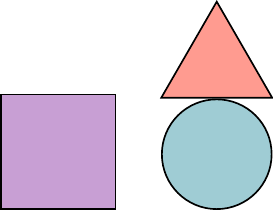}
    }\vspace{\abovecaptionskip}
    \subcaptionbox{Base frame}[0.3\linewidth]{
        \centering
        \includegraphics[width=0.75\linewidth]{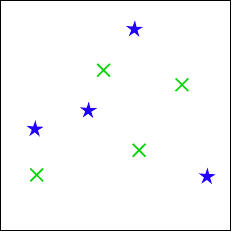}
    }\hfill
    \subcaptionbox{Circle frame}[0.3\linewidth]{
        \centering
        \includegraphics[width=0.75\linewidth]{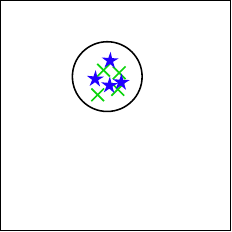}
    }\hfill
    \subcaptionbox{Square frame}[0.3\linewidth]{
        \centering
        \includegraphics[width=0.75\linewidth]{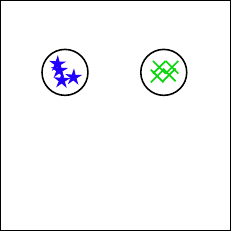}
    }
    \caption{The two actions of placing the triangle (figures \emph{a} and \emph{b}) are represented in three frames (figures \emph{c}-\emph{d}) and clustered. The clusters in \emph{c} represent the two contexts.}
    \label{fig:context_clustering}
\end{figure}

When solving tasks, both the context and the reference system of action is of importance. By context we mean how one action assumes different characteristics with respect to the particular instance in which the action is performed: for example, a place action that happens in the frame of an object \textit{A} but \textit{to the left} of an object \textit{B} is considered different from the same action if it happens \textit{to the right} of \textit{B} (Figure \ref{fig:context_clustering} - this is particularly important to solve tasks like the tower of Hanoi and it will be explained in detail in Section~\ref{sec:clustering}). Thus, our method relies on a reference frame inference algorithm using a clustering approach. The idea is that the same position in different demonstrations will have low variance and lie close together if represented in an adequate frame. Note that the context inference is also achieved with the clustering algorithm, as we do not include semantics for \textit{left, right, on, under, in}. In~\cite{niekum_learning_2015}, the authors represent the endpoint of a robot skill in different reference frames and identify clusters by using a threshold on the euclidean distance between the points. Similarly,~\cite{ekvall_robot_2008} infers reference frames by identifying clusters using K-means. The method in~\cite{dong_learning_2012} identifies clusters by fitting Gaussians and returning the reference frame with the lowest variance.

\section{Proposed Method} \label{sec:method}

\begin{figure}[tbp]
    \centering
    \resizebox{\linewidth}{!}{
        \input{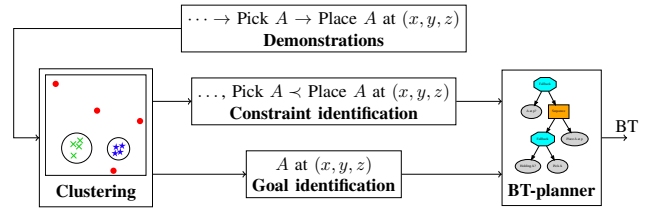}
    }
    \caption{Outline of the proposed method.}
    \label{fig:outline}
\end{figure}



At a high level, our proposed algorithm learns BTs from demonstrations in four steps, see Figure~\ref{fig:outline}. Human demonstrations are clustered to infer the context of each action and similarities between them, and then to infer task constraints and goal conditions, which are finally fed to a planner that builds the BT. By \textit{task constraints} we define sequences of actions whose relative order is important in the solving process. For example, if the robot has to place two items in a box, the order in which the items are placed might not be important, but it is important that an item is picked before it is placed. By \textit{goal conditions} we define the states in which the robot and the objects in the scene must be for the task to be considered completed.
In this section we detail all components of the method.
We restrict the work to consider only manipulation tasks, and by consequence we take examples from this domain. We argue that the validity of the algorithm is not limited by this choice but we leave the extension to mobile manipulation tasks and to other actions to future work.

\subsection{Demonstrating a task}

The teaching method is kinesthetic and there are three actions available: a \textit{Pick} action will close the robot grippers around the target object and a \textit{Place} or \textit{Drop} action will open the grippers, releasing the object. Thus, a demonstration is performed by dragging the robot end-effector to the desired pose and by selecting one of the available actions in a simple GUI. For all actions, the pose of the end-effector is recorded as the target pose for that action and in the reference frame of every object in the scene that matters for the task. The relevant objects, must be defined prior to the beginning of the demonstration.

\subsection{Actions and Conditions}

\begin{table}[tbp]
\centering
\caption{Available actions with their pre- and post-conditions. A ${}^*$ indicates that the action can be demonstrated, i.e. that is an action that the user can choose. The other actions are lower level and only available to the BT. The parameter $o$ represents an object while $\bm{p}$ is a position.}
\label{tab:actions}
\begin{tabular}{@{}llll@{}}
\toprule
\textbf{Action}                         & \textbf{Pre-conditions}          & \textbf{Post-conditions}           \\ \midrule
\textsc{Pick($o$)}${}^*$                & \textsc{Gripper($open$)}        & \textsc{Gripper(closed)}          \\
                                        &                                 & \textsc{InGripper($o$)}           \\
\textsc{Place($o$, $\bm{p}$)}${}^*$     & \textsc{InGripper($o$)}         & \textsc{ObjectAt($o$, $\bm{p}$)}  \\
                                        &                                 & \textsc{Gripper($open$)}          \\
                                        &                                 & \textsc{InGripper($none$)}        \\
\textsc{Drop($o$, $\bm{p}$)}${}^*$      & \textsc{InGripper($o$)}         & \textsc{ObjectAt$^\dagger$($o$, $\bm{p}$)}  \\
                                        &                                 & \textsc{Gripper($open$)}          \\
                                        &                                 & \textsc{InGripper($none$)}        \\
\textsc{SetGripper($open$)}             &                                 & \textsc{Gripper($open$)}          \\
                                        &                                 & \textsc{InGripper($none$)}        \\
\textsc{SetGripper($closed$)}           &                                 & \textsc{Gripper($closed$)}        \\ \bottomrule
\end{tabular}
\end{table}

\begin{table}[tbp]
\centering
\caption{List of incompatible conditions with the compatibility predicate: the conditions in every couple cannot hold at the same time if the predicate is satisfied.}
\label{tab:incompatible}
\begin{tabular}{@{}lll@{}}
\toprule
\multicolumn{2}{l}{\textbf{Condition Couples}}                             & \textbf{Incompatible if}                 \\ \midrule
\textsc{InGripper($o_1$)}          & \textsc{InGripper($o_2$)}             & $o_1 \neq o_2$                           \\
\textsc{Gripper($s_1$)}            & \textsc{Gripper($s_2$)}               & $s_1 \neq s_2$                           \\
\textsc{ObjectAt($o_1, \bm{p}_1$)} & \textsc{ObjectAt($o_2, \bm{p}_2$)} & $o_1 = o_2 \land \bm{p}_1 \neq \bm{p}_2$ \\
\textsc{ObjectAt$^\dag$($o_1, \bm{p}_1$)} & \textsc{ObjectAt$^\dag$($o_2, \bm{p}_2)$} & $o_1 = o_2 \land \bm{p}_1 \neq \bm{p}_2$ \\
\textsc{ObjectAt($o_1, \bm{p}_1$)} & \textsc{ObjectAt$^\dag$($o_2, \bm{p}_2)$} & $o_1 = o_2 \land \bm{p}_1 \neq \bm{p}_2$ \\
\textsc{GripperState($open$)}      & \textsc{InGripper($o$)}            & $o \neq none$                            \\ \bottomrule
\end{tabular}
\end{table}

To perform the tasks, the robot has access to the actions in Table~\ref{tab:actions}, listed with pre- and post-conditions. An action $A$ is formally defined as a tuple $A = (T, \bm{x}, \bm{p}, \bm{o}, F)$ where $T$ is the type (i.e. the action label), $\bm{x}$ is some action-specific parameter vector, $\bm{p}$ is the target position, $\bm{o}$ is the target orientation, and $F$ is the reference frame or coordinate system where the action occurs.

Each action $A$ is further given a set of pre-conditions $\mathcal{C}^A_{pre}$ that must be true for the action to succeed and a set of post-conditions $\mathcal{C}^A_{post}$ that are made true by executing the action. Let $\mathfrak{C}$ be the set of all conditions that appear as a pre- or post-condition of any action. Then, let the compatibility function
\begin{equation*}
    \mbox{comp}: \mathfrak{C}\times\mathfrak{C} \to \{true, false\}
\end{equation*}
take two conditions and return $true$ if they can be true at the same time and $false$ otherwise. Note that for any two conditions $c_1,c_2 \in \mathfrak{C}$ we have
\begin{align*}
    \mbox{comp}(c_1, c_2) &= \mbox{comp}(c_2, c_1) \\
    \mbox{comp}(c_1, c_1) &= true
\end{align*}
Incompatibility between conditions is detailed in Table \ref{tab:incompatible}.

The difference between \textit{Place} and \textit{Drop} lies in the accuracy with which the object is placed. In \textit{Place}, the post-condition is satisfied if the target object lies in a sphere of radius $5$~cm centered in the goal position of the action, while in \textit{Drop} a cylinder of radius $10$~cm is used (this is marked by a $^\dagger$ in the post-condition name). This matters since the target pose of the action is recorded in the end-effector. When dropping from a height, the recorded position will lie above where the object lands and the object may also bounce once it hits the ground. Thus, a \textit{Drop} action is chosen when the goal pose for the object doesn't need to be precise, e.g. when dropping objects in a trash bin.

\subsection{Behavior Tree synthesis}

The BT is synthesised using the planner proposed in~\cite{colledanchise_towards_2019}, leveraging the idea of backchaining. Note that unlike~\cite{colledanchise_towards_2019}, we run the planner in a simulator, as we don't want the robot to stop it's execution to re-plan in case of unmet post-conditions, and because it is preferable to have the full tree available before running it on a real robot. For this reason, the simulator used for the planner is set to cause the BT leaves to fail once, thus triggering the planner to expand the BT. In this, there is a trade-off to make between the tree size and the reactivity to unforeseen events when running on the physical environment.

Furthermore, we propose to give each action a cost. The cost could for example be execution time or any arbitrary ranking. If there are multiple actions that can achieve the same post-condition, only the one with lowest cost is added to the tree. If the goal is to have the gripper open, this avoids using the action \textsc{Place} instead of the more appropriate \textsc{SetGripper}($open$).

\subsection{Goal and constraints identification}

The algorithm to infer task constraints is based on the work proposed in~\cite{ekvall_robot_2008}, where constraints are identified by observing the order in which actions appear in the demonstrations and adding each pair of ordered actions to the list of constraints. We argue that it is not the order of the actions that is important, but rather the effects of one action to the next. In our method, the list of constraints generated by the order of demonstrated action is translated into pre-conditions that must be fulfilled before executing an action.

Conflicting constraints (for example action $A$ being performed after $B$ in one demonstration but before $B$ in another) are removed as they are interpreted as an aspect of the task that is not relevant and is left for the planner to solve. At the same time, candidate post-conditions of an action are checked for compatibility against the pre-condition of the following action and discarded if conflicts are generated.

Since we assume that by the end of the demonstration the goal is achieved, goal conditions are inferred by traversing demonstrations from end to start and recursively adding the actions' post-conditions. Demonstrations with the same goal conditions are grouped together and the final goal is satisfied if any of the groups' goal conditions are satisfied. Constraints are inferred for each group in isolation and the goals for each group are sent to the planner separately. The planned trees achieving these goal conditions are combined under a fallback node.

\subsection{Clustering of demonstrated actions} \label{sec:clustering}

Due to human errors and measurement noise, demonstrations of the same task might be inconsistent in the order of executed actions and in target positions of the objects. Different actions might also be executed in different reference frames. Thus, equivalent actions across demonstrations have to be identified and their reference frame inferred. An unsupervised clustering algorithm inspired by~\cite{niekum_learning_2015, ekvall_robot_2008, dong_learning_2012} is used to detect similarities between demonstrations. When representing the target position of the demonstrated actions in a reference frame $F$, similar actions that are executed in $F$ will lie close together and form clusters. The algorithm is extended here to not only use the clustering to infer reference frames, but also to detect context and action equivalence.

We define two actions $A$ and $B$ to be equal if all elements of the corresponding tuples are equal, except orientation. We assume that the position is important and different orientations are just variations of the same action. Note that we make this simplification as we deal with symmetric objects; it is not generally true.

Assuming $T$, $\bm{x}$, and $n$ are known, the clustering only has to be done among actions where these are equal since the actions cannot be equal otherwise. Once the candidate actions have been identified, their target positions are represented in each of the candidate frames. We detect clusters with the DBSCAN algorithm~\cite{ester_density-based_1996} as implemented in scikit-learn\footnote{\url{https://scikit-learn.org/stable/modules/generated/skleLarn.cluster.DBSCAN.html}}~\cite{pedregosa_scikit-learn_2011}, but any algorithm that does not require a-priori knowledge of the number of clusters and can reject outliers can be used. A score is given to each cluster according to
\begin{equation*}
    \mbox{score}(cluster) =
    \begin{cases}
        \frac{|cluster|}{r} \quad &|cluster| > 1 \\
        -\infty \quad &|cluster| = 1
    \end{cases}
\end{equation*}
where $|cluster|$ is the number of points in the cluster and $r$ is the maximum distance from a point in the cluster to the mean. The score function assigns a high score to large and dense clusters. If an action is part of multiple clusters, the cluster with the highest score is used. Actions that do not belong to any cluster are considered to not be equivalent to any other action and occurring in a default frame, typically map or robot base frame.

If an action potentially belongs to multiple clusters, we can infer the context in which the action is performed. It is desirable to detect when an action occurs in different contexts and treat each context as a separate action even if they have the same type, parameters, and target position, and occur in the same reference frame. Situations can arise where the same action $A$ is performed at two separate points in a demonstration. The constraint identification step would then notice that all actions between the two occurrences happen both before and after $A$. The result is contradicting constraints that are removed. All information about what happened between those two occurrences is thus lost. To solve this issue, the two best clusters will be kept if their score is above a threshold dependent on the number of demonstrations. 

To sum-up, the proposed method is capable of building a BT solving a task that has been previously demonstrated. In light of what explained, it is important to point out that for a given task, a single demonstration is enough for the method to generate a BT solving it. However, in such a case, the robot would just repeat what the human did. Therefore, the robot would be able to solve the task only if its starting conditions match the ones in the demonstration. If we want the robot to generalize, i.e. learn task constraints and infer the actions' reference frames, multiple demonstrations are required. For example, the clustering algorithm requires at least three samples to generate a cluster and infer another reference frame than the one in the robot base.

\section{Experiments and Results} \label{sec:experiments}

In this section we consider various manipulation tasks which challenge the learning algorithm in different ways. For the experiments from \textit{a)} to \textit{d)}, the demonstrations are performed by an user familiar with the system. The experiments were carried out using an ABB YuMi robot with an Azure Kinect camera for object recognition using Aruco markers. The code repository is made publicly available\footnote{\url{https://github.com/matiov/learn-BTs-from-demo}}.
The ABB YuMi robot is put in \textit{LeadThrough} mode (this mode switches off the motor brakes in the robot arm and it gravity-balances it) that allows to guide the robot end-effector to the desired position, then the user has to input the action type to perform. 

To demonstrate how sensitive the system is to varying demonstrations, in experiment \textit{e)} we asked non-expert users to solve a task, to gain some insight on the exploitability of the framework in an industrial scenario.

\begin{figure}[tbp]
    \centering
    \includegraphics[width=\linewidth]{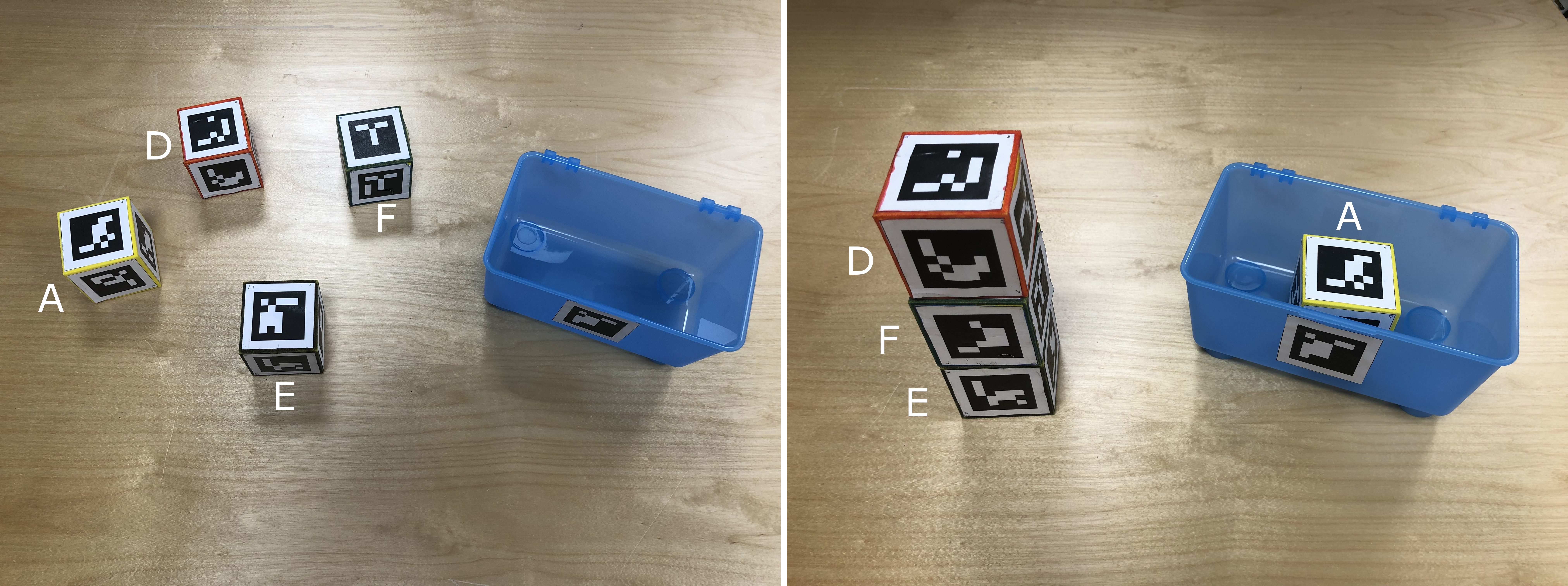}
    \caption{Initial (left) and goal (right) configurations for the task demonstrated by non-experts.}
    \label{fig:nonexpert_task}
\end{figure}

\paragraph{Object in box} \label{exp:box}
The goal of this task is to put a cube in a box (as for cube \textit{A} in Figure~\ref{fig:nonexpert_task}). From any starting condition, the demonstration is realized by picking the cube and then by dropping it in the box. 

\textbf{Results:} The algorithm is able to learn a BT after only one demonstration. However, three demonstrations are required to solve the task flawlessly. In this case it learns that dropping the cube must happen in the box frame to make the learned BT (reported in Figure~\ref{fig:example_bt}) robust to changes in the box position. The task can be solved by choosing both the \textit{Place} or \textit{Drop} actions. However, if the former is preferred, then the user has to be careful not to open the grippers above the box, but inside it. If \textit{Place} is used instead of \textit{Drop} when the robot end-effector is above the box, the cube could roll or bounce inside the box, outside the more restrictive tolerance of the action. Because of the reactivity of BTs, the robot would attempt the picking again, until the task succeeds or the picking action fails due to collision with the robot and the box. The BT generated for this task had 16 nodes.

\begin{figure}[tbp]
    \centering
    \includegraphics[width=\linewidth]{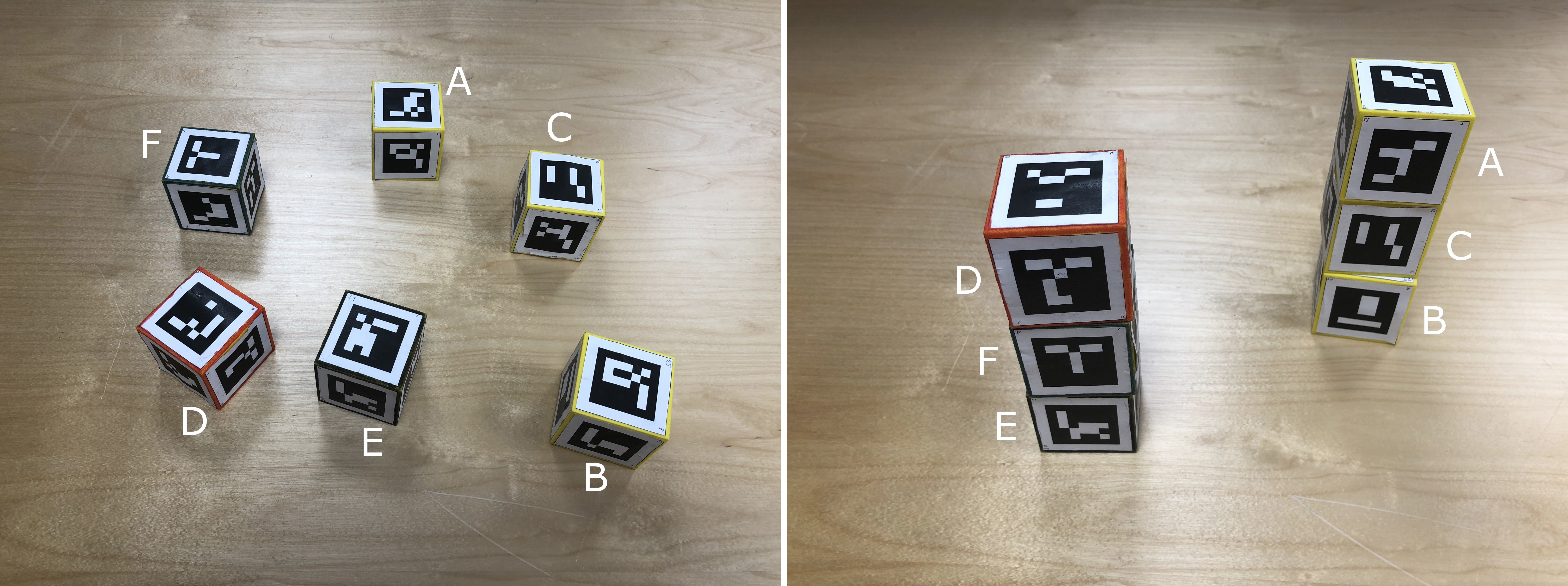}
    \caption{Initial (left) and goal (right) configurations for the towers stacking task.}
    \label{fig:towers_task}
\end{figure}

\begin{figure*}[tbp]
    \centering
    \includegraphics[width=.9\linewidth]{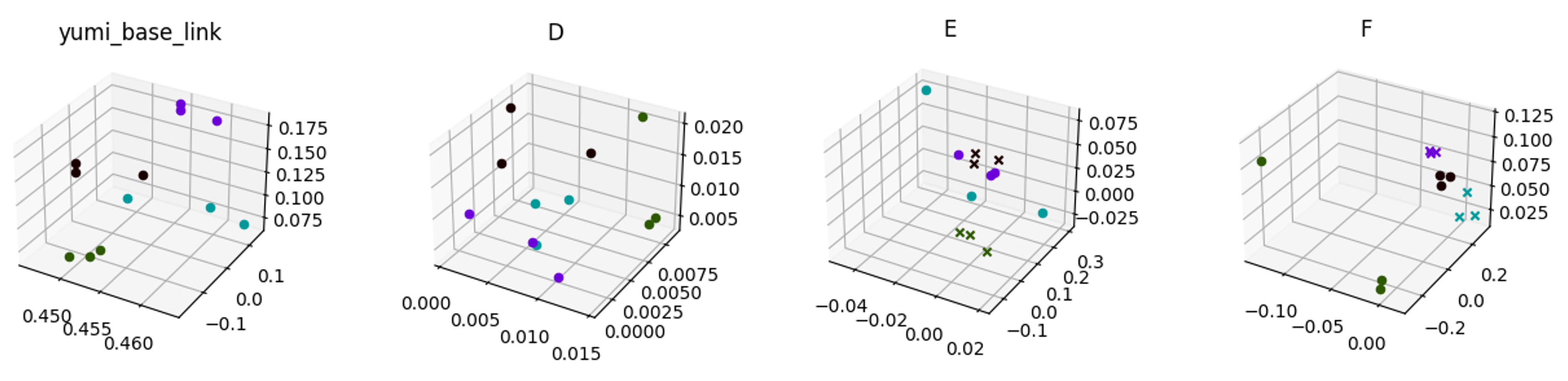}
    \caption{Clustering step for the \textit{Place D} action in the tower of Hanoi task. In our method, since the action \textit{Place D on top of E} happens twice but in two different contexts, it is detected as being two different actions. We can see that the purple and black samples lie close to each other both in the reference frame \textit{E} and \textit{F}. However, the reference frame \textit{E} is used for the action \textit{Place D} in the context of Figure~\ref{fig:hanoi_context1}, while the reference frame \textit{F} is used in the context of Figure~\ref{fig:hanoi_context2} (being the choice labeled with `\texttt{x}'). Samples of other colors represent the action \textit{Place D} in other steps of the solving process.}
    \label{fig:hanoi_cluster}
\end{figure*}

\begin{figure}
     \centering
     \begin{subfigure}[b]{0.45\textwidth}
         \centering
         \includegraphics[width=\textwidth]{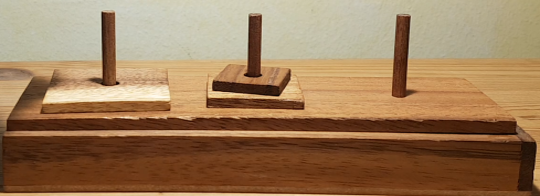}
         \caption{In this step of the solving process, the smallest disk \textit{D} is placed on top of the medium disk \textit{E}, while it lays on the medium pole.}
         \label{fig:hanoi_context1}
     \end{subfigure}
     \hfill
     \begin{subfigure}[b]{0.45\textwidth}
         \centering
         \includegraphics[width=\textwidth]{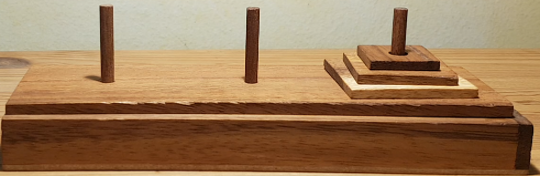}
         \caption{In the final step of the solving process, the smallest disk \textit{D} is placed on top of the medium disk \textit{E}, while it is in turn placed on top of the largest disk \textit{F}.}
         \label{fig:hanoi_context2}
     \end{subfigure}
        \caption{In both these two steps of the tower of Hanoi puzzle, the smallest disk \textit{D} has to be placed on top of the medium disk \textit{E}. However, this action happens in two different contexts as the relative position of all three disks is different in the two steps and therefore it has to be taken into account.}
        \label{fig:hanoi_context}
\end{figure}

\paragraph{Towers stacking}
The goal of this task is to stack the cubes to form towers as showed in Figure~\ref{fig:towers_task}. The demonstration is realized by picking one cube at a time and fine placing it in the desired position. 

\textbf{Results:} The algorithm is robust to variations in the demonstrations, where the base of the tower is moved before stacking the remaining cubes or where the order of the cubes is altered from a demonstration to another. In such cases, the robot would learn different possible configurations for the towers and grow a BT for each configuration, thus affecting the size of the final tree. If multiple demonstrations are provided, the placing actions are learned to be in the frame of the base of the tower. The BT that builds the two towers featured 101 nodes.

\paragraph{Towers of Hanoi}
As shown in our previous work~\cite{styrud_combining_2022}, this is a challenging task for most basic planners and thus motivates the needs of using learning approaches. To solve this task one has to stack disks with decreasing radius and the same action of placing a smaller disc on top of a larger one can happen at different steps in the solving process. We reproduce the task by moving a stack of three cubes from a position to another (see accompanying video). 

\textbf{Results:} Solving the task requires knowledge of the context in which actions happen, learning frames is not enough. Thus, our clustering step is necessary to solve the ambiguity of having the same placing action happening in different contexts (see Figures~\ref{fig:hanoi_cluster} and Figure~\ref{fig:hanoi_context} for a detailed analysis, where cubes \textit{D}, \textit{E}, \textit{F} are used for the task). Without the context inference, the contradictory constraints generated by the repeated actions, cause the planner to get stuck in the local minima of Figure~\ref{fig:hanoi_context1}. The BT generated to solve this task had 113 nodes.

\paragraph{Kitting}

\begin{figure}[tbp]
    \centering
    \includegraphics[width=\linewidth]{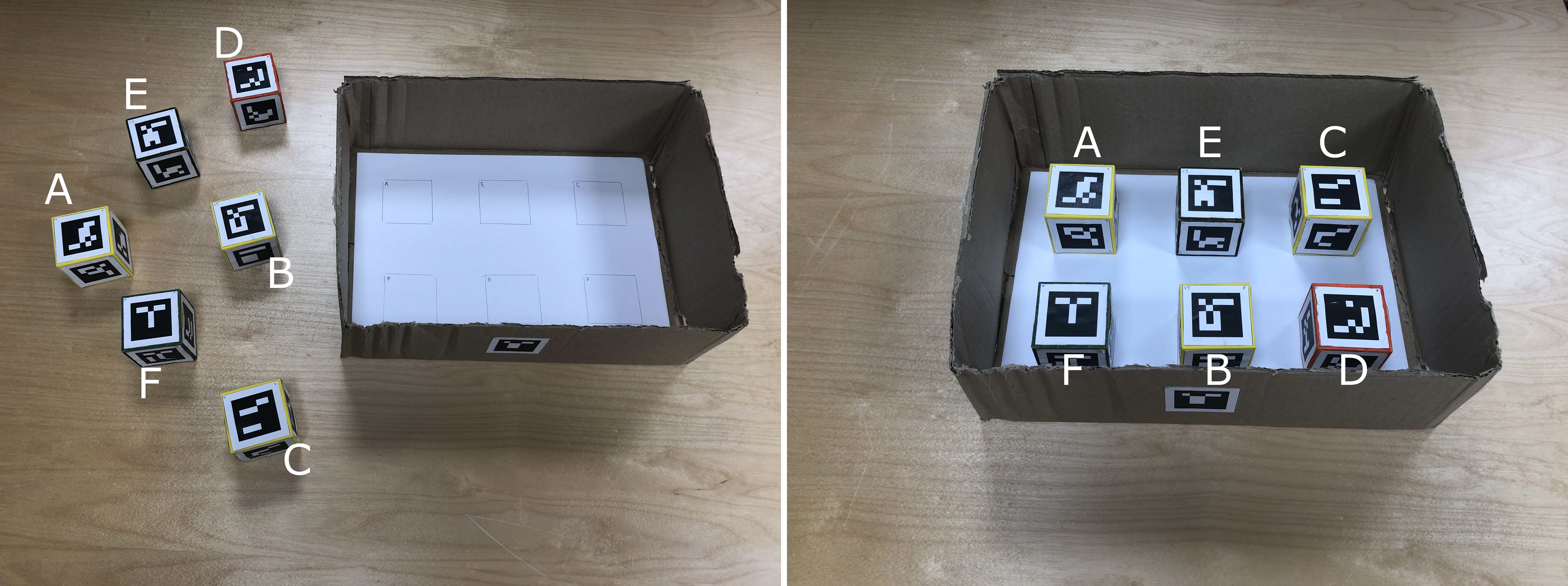}
    \caption{Initial (left) and goal (right) configurations for the kitting task.}
    \label{fig:kitting_task}
\end{figure}

In this task the goal is to place the cubes in specific positions inside a bigger box as in Figure~\ref{fig:kitting_task}, to mimic a kitting task. The task is demonstrated by picking one object at a time and fine placing it inside the box in the marked area. Multiple demonstrations are provided where the order in which objects are picked and placed varies. 

\textbf{Results:} Since what matters is the place pose for each object, not the order in which the objects are placed, there are many possible ways of solving the task (the demonstrations covering only a few) and the risk is that non-relevant constraints are inferred (of type place object \textit{A} before \textit{B}). This makes the learned tree, with 603 nodes, larger than an optimal solution but does not compromise solving the task.

\paragraph{Non-expert demonstrations}

We asked employees at ABB Corporate Research to demonstrate a task featuring a tower with three stacked cubes and an object in a box. Some of the users were familiar with the \textit{LeadThrough} mode of the robot but none with the learning algorithm and were thus instructed to dispose the cubes in the goal configuration starting from a random initial condition (Figure~\ref{fig:nonexpert_task}). Each user demonstrated the same task three times for the same reasons as in paragraph~\textit{a)}. The goal of this experiment is to test if non-expert users can indeed exploit the learning framework and how their demonstrations affect the learned BT. Note that designing a user-friendly GUI was not within the scope of the work, so we did not ask the users to provide subjective feedback on the system usability.
Some users have also tested the system to demonstrate tasks of their own choice, e.g. stacking cubes in pyramid-like configurations, showing that the learning algorithm is also generalizable (within the limit in the number of tasks that can be designed with four cubes and a box).
Due to the precision of the vision system affecting the results, the reported numbers for the estimates on the success rate should not be taken as accurate performance measures but rather an indication.
To this extent, we aim to analyze the performance of the learned BTs and the differences between BTs in terms of number of nodes. Finally, we study the behavior of the learning algorithm when it is given demonstrations from several different users.
Out of the 14 users that tested the system, 7 made the optimal choice for the actions to use in the different parts of the task, while 4 made a less optimal choice leading the robot to fail the task in some cases. For the last user there has been an error in the frame inference, leading to an unsuccessful tree. We removed demonstrations from 2 users who misunderstood the task, and tried to deliberately trick the system.

\begin{figure}[tbp]
    \centering
    \includegraphics[width=\linewidth]{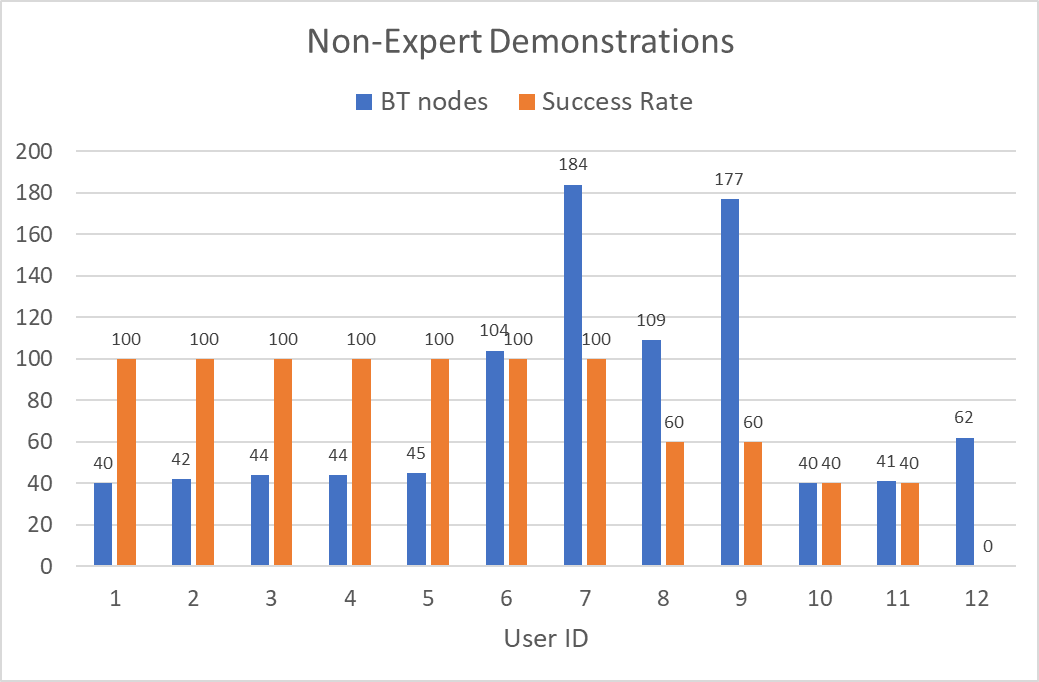}
    \caption{Number of nodes and success rate of the BTs generated from non-expert demonstrations, before correcting action types. The data has been sorted for clarity.}
    \label{fig:nonexpert_chart}
\end{figure}

\textbf{Results:} The results of this experiment are plotted in Figure \ref{fig:nonexpert_chart}, where the success rate is computed over 5 runs. 
When the users are consistent in all the three demonstrations, a success rate of $100\%$ is achieved and the size of the tree is limited (users~1 to~5). Moreover, in these demonstrations the two subtasks of building the tower and placing the cube in the box are solved in sequence. Minor variations, e.g. the subtasks are not well separated (i.e. two cubes stacked, then cube placed in the box and finally tower completed - user~6), or the base of the tower (i.e. the E~cube in Figure \ref{fig:nonexpert_task}) is moved before stacking the other cubes (user~7), result in an increased size for the learned tree, without compromising the success rate.
The success rate is lower for users~8 to 11. This is caused by their non-optimal choice of the placing actions. Users~8, 9 and 11 chose \textit{Drop} (instead of \textit{Place}) to stack the cubes, with the tree achieving the more relaxed condition for the goal pose of the tower cubes if they happened to be close to each other but not stacked. With this choice it is still possible to build a successful tree completing the task, but if for some reasons the tower is undone and the cubes happen to be close to each other, the tree will not recover because the goal conditions for the cubes would still be satisfied. User~10 chose \textit{Place} (instead of \textit{Drop}) to put the cube in the box, thus the system learns the goal pose for the cube with too tight tolerances that are not always achieved. Letting these users manually change the action labels after testing the BT fixes the problem, building successful BTs.
For user~12 the algorithm learns to build the tower in the frame of cube~A, thus when the cube is moved the robot tries to move the base of the tower accordingly, resulting in a collision, as it has not learned to undo the tower. This clustering error can be corrected by combining the demonstrations from user~12 with e.g. users~6 and 7, and by increasing the number of samples to form a cluster, so that the algorithm is forced to use demonstrations from more than one user.

We can conclude that the learning algorithm behaves as good as the data it is provided, meaning that if the task is demonstrated as intended, but still allowing for minor variations, then the performance is satisfactory. It deteriorates if the user makes non-optimal choices in the actions or if a user deliberately intends to trick the system. We can assume that in an industrial scenario users have all the interests in making the demonstrations as precise as possible, since they need to collaborate with the robot towards a common goal.


\section{Conclusions and Future Work} \label{sec:conclusion}

In this paper we proposed a method that allows to build BTs from demonstration and infer important features of the task to solve. 
A natural extension of this work would be to extend the set of tasks and behaviors to mobile manipulation applications. In such a case, the navigation part would be taught by teleoperation. 
The proposed method is tested on toy problems with the intent to mimic assembly tasks, where it is important to learn relative positions between objects. The method works with any perception algorithm that can estimate the pose of the objects in the scene by creating a reference frame for every object. We use marker detection for its simplicity, aware that it may not be a viable solution for industrial scenarios.
We would want to use common objects instead of cubes and object detection algorithms for the perception. Removing the markers might prevent the robot from recognising objects univocally: if there are two objects of the same category, the robot won't be able to tell one from the other. To solve this issue we exploit current Human Robot Interaction (HRI) frameworks (leveraging the camera depth information, as in~\cite{dogan_followup_2022} to give the robot the capability to interact with the operator to disambiguate the objects~\cite{iovino_interactive_2022}.

BTs learned from demonstration could also be used to seed a Genetic Programming (GP) algorithm~\cite{iovino_learning_2020, styrud_combining_2022}, for example in a context in which the human operator demonstrates a subtask of a more complex task and finding the whole solution is left to the GP.

At its current state, the method would benefit from some post-processing features, e.g to allow the user to shrink and expand the BT while visualizing it. If we consider the first subtree rooted with a \textit{Fallback} node in the BT from Figure~\ref{fig:example_bt}, it could be visually shrank to a box with label \textit{`Pick A and Place at p1 in box'}. The user could then click on said box and see the expanded subtree. Another useful feature could be to automatically combine similar subtrees to optimize the full BT size. This is something that a GP algorithm could do, but attention must be paid not to sacrifice human readability in the optimization process.

Even if multiple demonstrations allow the robot to generalize the task, demonstrating again the same task is a tedious operation. Moreover, at its current state, all information is lost once the tree is learned. By this, we mean that if operators want to teach the robot a task that logically builds on previously learned tasks, they have to perform the whole demonstration again. As an example, say that we previously performed the demonstrations for experiment \textit{a)} but we now want to drop a second cube inside the same box. In the current state, the operator (without manually modifying the BT) would need to perform the whole demonstration again, for three times. Solutions to both problems - avoiding repeating a same demonstration multiple times and exploiting previously learned trees - are left as future work.



\bibliographystyle{IEEEtran}
\bibliography{references}

\end{document}